# Robust event-stream pattern tracking based on correlative filter

**Hongmin Li[1]** · **Luping Shi[2]\***

## Abstract

Object tracking based on retina-inspired and event-based dynamic vision sensor (DVS) is challenging for the noise events, rapid change of event-stream shape, chaos of complex background textures, and occlusion. To address these challenges, this paper presents a robust event-stream pattern tracking method based on correlative filter mechanism. In the proposed method, rate coding is used to encode the event-stream object in each segment. Feature representations from hierarchical convolutional layers of a deep convolutional neural network (CNN) are used to represent the appearance of the rate encoded event-stream object. The results prove that our method not only achieves good tracking performance in many complicated scenes with noise events, complex background textures, occlusion, and intersected trajectories, but also is robust to variable scale, variable pose, and non-rigid deformations. In addition, this correlative filter based event-stream tracking has the advantage of high speed. The proposed approach will promote the potential applications of these event-based vision sensors in self-driving, robots and many other high-speed scenes.

[1] H. Li

Center for Brain-Inspired Computing Research (CBICR), Optical Memory National Engineering Research Center, Department of Precision Instrument, Tsinghua University, Beijing 100084, China Tel.: +86 18810454635
Fax: +86 10 62788101
E-mail: li-hm14@mails.tsinghua.edu.cn

[2] L. Shi

\* Corresponding author

Center for Brain-Inspired Computing Research (CBICR), Optical Memory National Engineering Research Center, Department of Precision Instrument, Tsinghua University, Beijing 100084, China Tel.: +86 13910693300
Fax: +86 10 62788101

# 1 Introduction

Different from the traditional frame-based imager, the retina-inspired, dynamic vision sensor (DVS) converts the temporal contrast of the light intensity in visual scenes into spatiotemporal, sparse event streams. [2] [31] [17] This kind of event-based sensors eliminate the need to extract movement from sequences of image frames. The DVS sensors have the advantage of low information redundancy, high dynamic range and high speed in visual sensing, and have many potential applications in the  high-speed scenes.[22] [5] [23] [6] [7] [20] For example, DVS sensor has been used for the high-speed generation of intensity field [20][1]. In addition, a visual processing system based on the silicon retina has also the advantages of low computational complexity, low energy consumption. This kind of sensors are robust to varying intra-scene illumination for that the pixel is designed to be only sensitive to the local temporal contrast, rather than the absolute light intensity. The outputted events are represented in the form of Address-Event Representation (AER)[3]. AER is often utilized to model the biological systems, like the retina, using discrete time events to convey information, similar to the spike coded neural systems in living organisms.

Visual tracking has a wide range of applications in the fields of self driving, robot vision, trajectory analysis and so on.

 When an object is detected at a certain moment, it is often useful to track that object in subsequent recording. Many works of object tracking based on the DVS sensor have been reported. [16] [26] [24] [25] [28] [19] [30] [29] [11] [15] [32][18] Although some existing methods have achieved some satisfactory performance in several relatively simple scenes, however, event-stream object tracking is a challenging task due to the significant appearance variations caused by  the noise events, complex background textures, occlusion and randomness of event generating in the pixel circuit. Firstly, events stimulated by the contour and textures of the object are easy to be confused by events from the background textures. If the target event-stream pattern has the similar spatiotemporal shape with that of a background object, a tracker based on simple feature representation would be easily confused. Besides, the event-stream shape of the non-rigid object changes and deforms all the time, which demands a more discriminative feature representation. Figure 1 shows some successive reconstructed frames from several event-stream recordings. The ground-truth position of the target object is shown with a bounding box in each segment. From the pictures, we can see that the appearance of the target event-stream object changes obviously even between two adjoining segments, which demands a robust tracker for tracking the rapid changed appearance.

 This paper presents a robust event-stream pattern tracking method based on the correlative filter mechanism. Rate coding was used to encode the event streams of the target object. Hierarchical convolutional layers of a convolutional neural network (CNN) are used to extract the feature representation. The performance of the proposed method is evaluated on the DVS recordings of several complicated visual scenes. Among the recordings, three are captured by a DVS128 sensor (128×128 resolution)[31] by ourselves and the rest are from an event-stream tracking dataset [14] captured by a DAVIS sensor (240×180 resolution)[2]. The results prove that the proposed method can successfully track the specified objects in the visual scenes with noise events, complicated background textures, occlusion, and intersected trajectories.

## 2 Related Works

The event-stream pattern tracking methods can be classified into two categories. The first category is based on the event-driven mechanism in which each incoming event is processed and determined whether it belongs to the target object. In [19], M. Litzenberger et al. implemented a continuous clustering of AER events and tracking of clusters. Each new event was assigned to a cluster based on a distance criterion and was used to update the clusters weight and center position. In addition, point cloud method is also introduced to model the event-stream object. In [25], Z. Ni et al. proposed an iterative closest point based tracking method by providing a continuous and iterative estimation of the geometric transformation between the model and the events representing the tracked object. In [24], they developed a event-based microrobotic tracking system to track a microgripper position based on the iterative closest point algorithm. One drawback of the event-driven methods is that in the event-flow recording, there are a great number of noise events which make the tracker make a wrong inference. Adding a noise event filtering module to the tracking system would increase the computational complexity of the system, while many informative events would also be inevitably filtered. Besides, a great deal of events are generated within a short time period, which would be a limitation to the real-time application. Finally, although these event-based sensors are based on the event-driven nature, it is still a difficult task to recognize an object from a single event. The second category is based on the feature representation of the target object. In [33], the authors proposed a soft data association modeled with probabilities relying on grouping events into a model and computing optical flow after assigning events to the model. In [16], Xavier Lagorce et al. proposed an event-based multi-kernel algorithm, in which various kernels, such as Gaussian, Gabor, and arbitrary user-defined kernels were used to handle variations in position, scale and orientation. In [18], H. Li presented a compressive sensing based tracking method based on the rate coding of the event-stream object resulting in good performance on several simple scenes.

  The core of most modern tracking methods is a discriminative classifier to distinguish the target from the surrounding environment. Among many trackers, CF based methods has enjoyed great popularity due to the high computational efficiency with the use of fast Fourier transforms [9] [8] [13] [12] [21] [4] [13]. In [4], Bolme et al. trained the CF over luminance channel the first time for fast visual tracking, named MOSSE tracker. In [13], a kernelized correlation filter (KCF) is introduced to allow non-linear classification boundaries. In [9], color feature is used to improve the tracking performance. Besides, features from convolutional neural network (CNN) have achieved good performance in many computer tracking methods. In [8], the learned CNN features rather than hand-crafted features were used to improve the performance of CF based framework and the initial convolutional layer is demonstrate providing the best performance for tracking. In [21], multiple correlation filters on hierarchical convolutional layers show good tracking performance in many visual scenes. In this paper, hierarchical convolutional layers from rate coding of event-stream object are used for robust feature representation of the event-stream appearance.

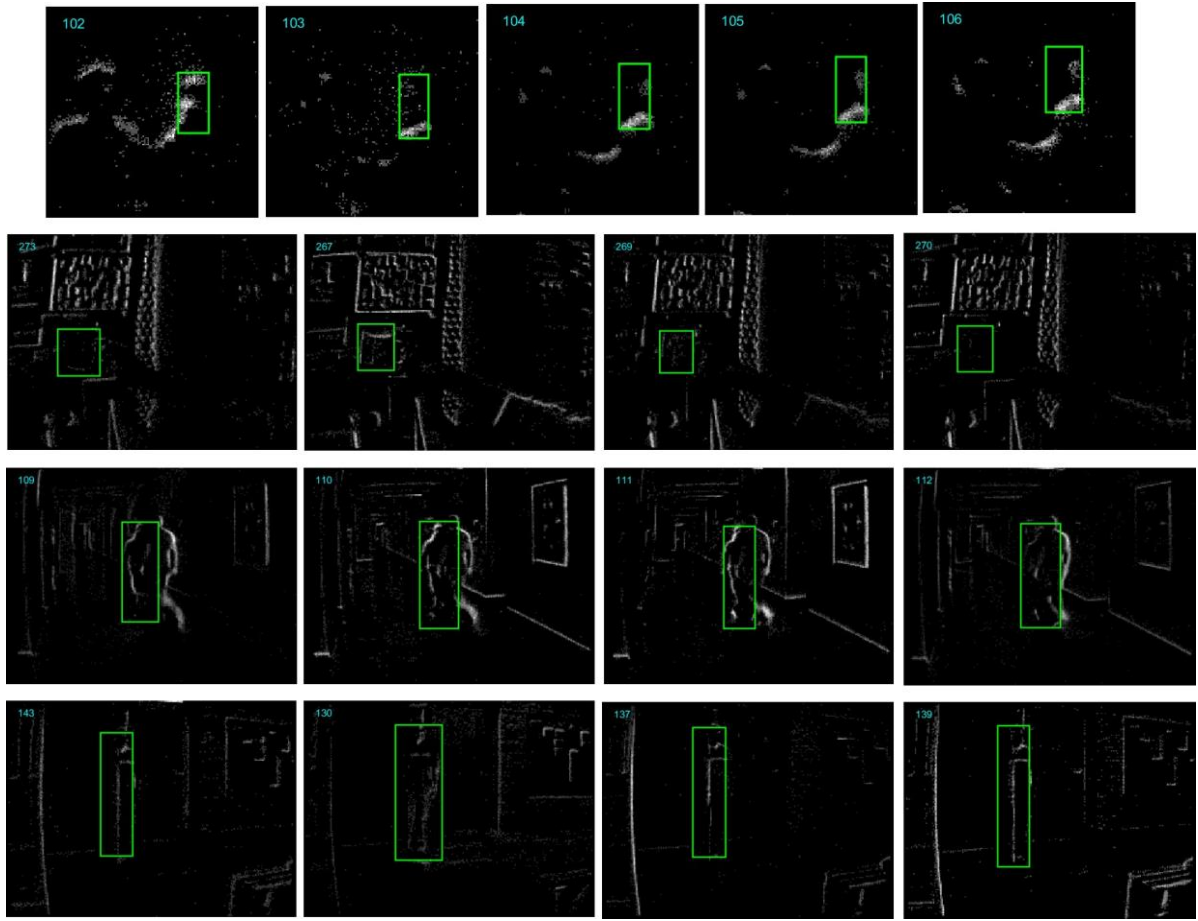

**Fig. 1** Some example reconstructed frames from several event-stream recordings (From top to bottom are the "Horse toy" from DVS128 sensor, "Vid B cup", "Vid J person floor", and "Vid E person part occluded" from DAVIS sensor). The appearance of the target object (in bounding box with green line) changes obviously for the noise events, background textures, and randomness of the pixel firing.

# 3 Methodology

## 3.1 Temporal Contrast Pixel

The pixel of the DVS sensor is a type of temporal contrast pixel which only responses to the temporal contrast of the light intensity in the scene and generates a temporal event whenever the brightness change exceeds a pre-defined threshold. Each event is a quadruple $(x,y,t,p)$, where $(x, y)$ denotes the positions of the pixel, $t$ denotes the timestamp, the polarity $p = 1$ denotes the increasing brightness and $p = -1$ denotes the decreasing brightness. The temporal contrast pixel has the advantage of high dynamic range because it needs not to respond to the absolute light intensity. The time stamp of each event has the temporal resolution of microsecond. Then the DVS sensor is capable to capture the high-speed dynamic scenes. Figure 2 shows the event generating principle of the temporal contrast pixel. The temporal contrast denoted as follows,

$$TCON = \frac{1}{I(t)}\frac{dI(T)}{dt} = \frac{d(ln(I(t)))}{dt} \quad (1)$$

where $I(t)$ is the photocurrent which is a function of time.

In this work, we use several event-stream recordings from both DVS128 and DAVIS sensors to evaluate the performance of the proposed method. The both two sensors are based on the same temporal contrast mechanism. As the name of the sensor shows, DVS128 has the spatial resolution of 128×128. DAVIS is a new retina-inspired event-based vison sensor with the spatial resolution of 240 × 180.

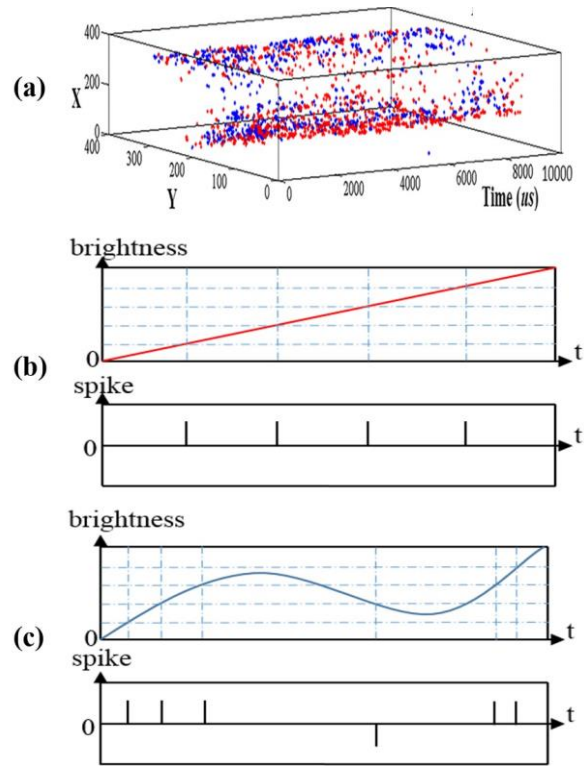

**Fig. 2** The event-generating principle of temporal contrast pixel. (a) Spatial-temporal event stream of a banana. Only the temporal contrast of the light intensity trigger the event output. (b) Linear temporal intensity change generates events with identical interval. (c) Nonlinear change generates events with varying intervals.

## 3.2 Motion Symbol Detection and Rate Coding

Although these event-based vision sensors are based on the event-driven nature, it is still a difficult task to recognize an object from a single event. In fact, it is the cluster of events that form the shape of the target object. In many works, event streams have been accumulated into multiple segments on which to extract feature for information processing [7][27], such as the event-stream object display in jAER open-source tool[10]. There exist two accumulating methods, i.e., hard events segmentation (HES) and soft events segmentation (SES). HES divides the event flow into segments using fixed time slices or fixed number of events. Different from HES, SES adaptively obtains the segments according to the statistical characteristics of the events based an event responding neuron, such as the leaky-integration-firing neuron.

For comparison, we segment the event streams into the same number of segments with the items of the ground-truth. Rate coding is employed to encode the visual information of the event-stream object. Intuitively, each pixel value is represented as the number of events generated by this pixel within the segment. From the view of the device, the temporal rate coding of each pixel is determined by a pre-defined threshold and the temporal intensity change *TCON* with a time window dt as follows,

$$Rate(t) = \frac{TCON}{\theta} = \frac{d(ln(I(t)))}{dt}\frac{1}{\theta} \qquad (2)$$

Within a time window, the physical meaning of the number of events of a pixel represents the frequency of which the temporal intensity change exceeds the pre-defined threshold. In rate coding, the serious temporal noise of the events is suppressed by ignoring the temporal information of the events in the segment.

### 3.3 Correlative Filter Framework

In this work, we use a CF tracking framework to overcome the challenging in the event-stream pattern tracking task. Generally, a CF tracker learns a discriminative classifier and finds the maximum value of the correlation response map as the estimation of the position of the target object. The resulting classifier is a correlation filter which is applied to the feature representation. Circular correlation is utilized in the CF tracking framework for efficiently train. In this work, we use the feature maps of the CNN as the mult-channel representation. For the l-th layer output of the neural network, the feature maps is denoted as $x^l$ of size $H \times W \times C$, where $H$, $W$ and $C$ denote the width, height, and the channel number, respectively. The correlation filter $f_l$ has the same size with the feature maps in the current frame $t$. In CF framework, the correlation filter is trained by solving a linear least-squares problem as follows, where $x^l_{h,w}$ demote the shifted sample. $h \in \{0,1,2,...H-1\}$, $w \in \{0,1,2,...W-1\}$. $y_{h,w}$ is the Gaussian function label, and where $\lambda$ is a regularization parameter $(\lambda \geq 0)$. $y_{h,w} = exp(\frac{-(h-H/2)^2+(w-W/2)^2}{2\sigma^2})$, where $\sigma$ is the kernel width. The minimization problem in (2) can be solved in each individual feature channel using fast Fourier transformation (FFT). We use the capital letters as the corresponding Fourier transform of the signal. The learned correlation filter in frequency domain on the $c$-th $c \in \{1,2,...,C\}$ channel is as follows,

Where the operation $\odot$ denotes the element-wise product, $Y$ is the Fourier transformation form of $y = \{y_{h,w} \mid h \in \{0,1,2,...H-1\}, w \in \{0,1,2,...W-1\}\}$, and the bar means complex conjugation. Given an rate coding patch in the next segment, the feature map on the $l$-th layer is denoted by $z^l$ and of size $H \times W \times C$. The $l$-th correlation response map of size $H \times W$ is calculated as follows, where the operation $\mathcal{L}^{-1}$ denotes the inverse FFT transform. The position of maximum value of the correlation response map $r^l$ is used as the estimation of the target location on the $l$-th convolutional layer.

---

[1] The model can be download from: $http : //www.robots.ox.ac.uk/\ albanie/models/ssd/vgg - vd - 16 - reduced.mat$

## 3.4 Representation based on Convolutional Neural Network

Instead of using hand-crafted features, the feature maps of a CNN network have demonstrated a discriminative representation in many tracking systems. In this work, multiple convolutional layers are combined to encode the changed appearance of the event-stream object. A VGG-Net-16 model [1]implemented in the MatConvNet library is used in our method for feature extraction. The used CNN model was trained on the large-scale ImageNet for image classification task. Differently, instead of resizing the size of the input rate coding maps to equal the size of the input layer of CNN (i.e., 224 × 224), in this work, we use the resulted model parameters of each layer to perform convolutional operation on the original input. As the rate coding map is single-channel, we simply set the three channels of the input layer all equal to the rate coding map. Because the pooling operation would reduce the spatial resolution with the increasing depth of convolutional layers, we didn't use the layers higher than the conv3_3 layer. In this work, we test some hierarchical composition of different convolutional layers for feature representation and found that the representation of composition of conv1 1, conv2 2 and conv3 3 achieved a satisfactory tracking performance.

# 4 Experimental Results and Discussion

In this section, a series of experiments on several event-stream recordings are presented. The metric used in this paper is the center location error measured with manually labeled ground truth data.. Section 4.1 introduces the event-stream tracking dataset on which we perform the tracking experiments. In Section 4.2, we evaluate the influence of two hyper-parameters. Section 4.3 presents the tracking speed over different layers of CNN. In Section 4.4, we compare the performance of the proposed method with some other event-stream tracking methods.

## 4.1 Event-stream Recordings

Eight event-stream recordings of which three were captured by a DVS128 sensor [31] by ourselves, and the rest are from an event-stream tracking dataset captured by DAVIS sensor [14] were employed in this work.

*Three DVS128 recordings* [1].Three event-stream recordings of three different scenes were captured with a DVS128 device [31]. These recordings have the same spatial resolution of 128 × 128. We divided each recording into multiple segments and in each segment, we label the position and the size of the target object with a bounding box. The first recording is a scene containing many digits captured by a moving DVS128 sensor. The task is to track the Dight3 in the event streams. The time range of this  recording is about 38.8 second. We divided the recording to 767 segments. The second recording is a moving human with a Horse toy in his hand. The task is to track the Horse toy in the event streams. This is a video of about 17.3 second and divided into 347 segments. The appearance of the toy changed quickly with rotation and deformation in the recording. The third recording is video of a moving human of about 17.1 second. This recording is divided in 343 segments. The task is to track the face of the human. The human face moved fast with rotation and deformation and the appearance of the event stream changed all the time, which make the task difficult at a high speed.

---

[1] The DVS128 recordings and the groundtruth, named "DVS recordings and groundtruth.zip" can be download from:*https : //figshare.com/s/70565903453eef7c3965*

*DAVIS recordings* [1]. Yuhuang Hu et al [14] proposed an event-stream tracking dataset with a DAViS240C neuromorphic vision sensor with the spatial resolution of 240× 180. In some tracking sequences of this tracking dataset, the target objects are still, or cannot be differentiated from the background. We chose five recordings from this dataset. We also re-labeled the bounding boxes of the target objects as the groundtruth because the original provided bounding boxes seem to have a little shift compared to the precise position of the target objects. The five recordings are "Sylvestr", "Vid B cup", "Vid C juice", "Vid E person part occluded", and "Vid J person floor", and are divided into 1344, 629, 404, 305 and 388 segments in the groundtruth, respectively. These recordings contain several challenges of complicated background textures, noise events, randomness of pixels in generating events, and occlusion.

*Challenges in each recording.* We list the challenges in each recordings as show in Table 1. Seven challenges are taken into consider, including the noise events, complicated background, occlusion, intersected trajectories, deformation, changing scale and changing pose. One or several challenges are contained in each recording.

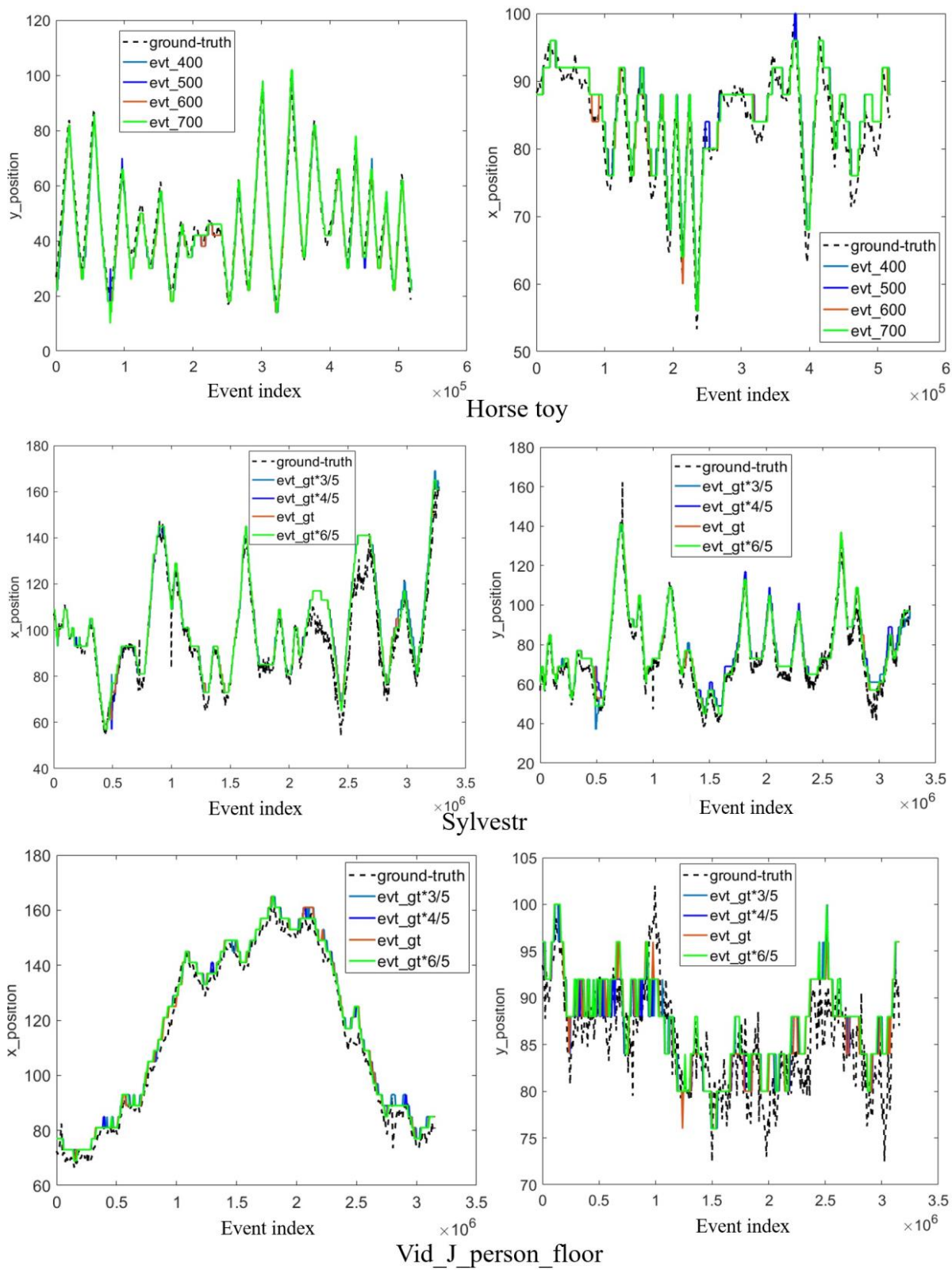

**Fig. 3** Tracking performance under different event number per time bin on three event-stream recordings (from top to bottom are "Horse toy", "Sylvestr", and "Vid J person floor" scenes, respectively). The x (left) and y

(right) positions of the center of the tracker over the time are displayed. We use the index of event to represent the time coordinate.

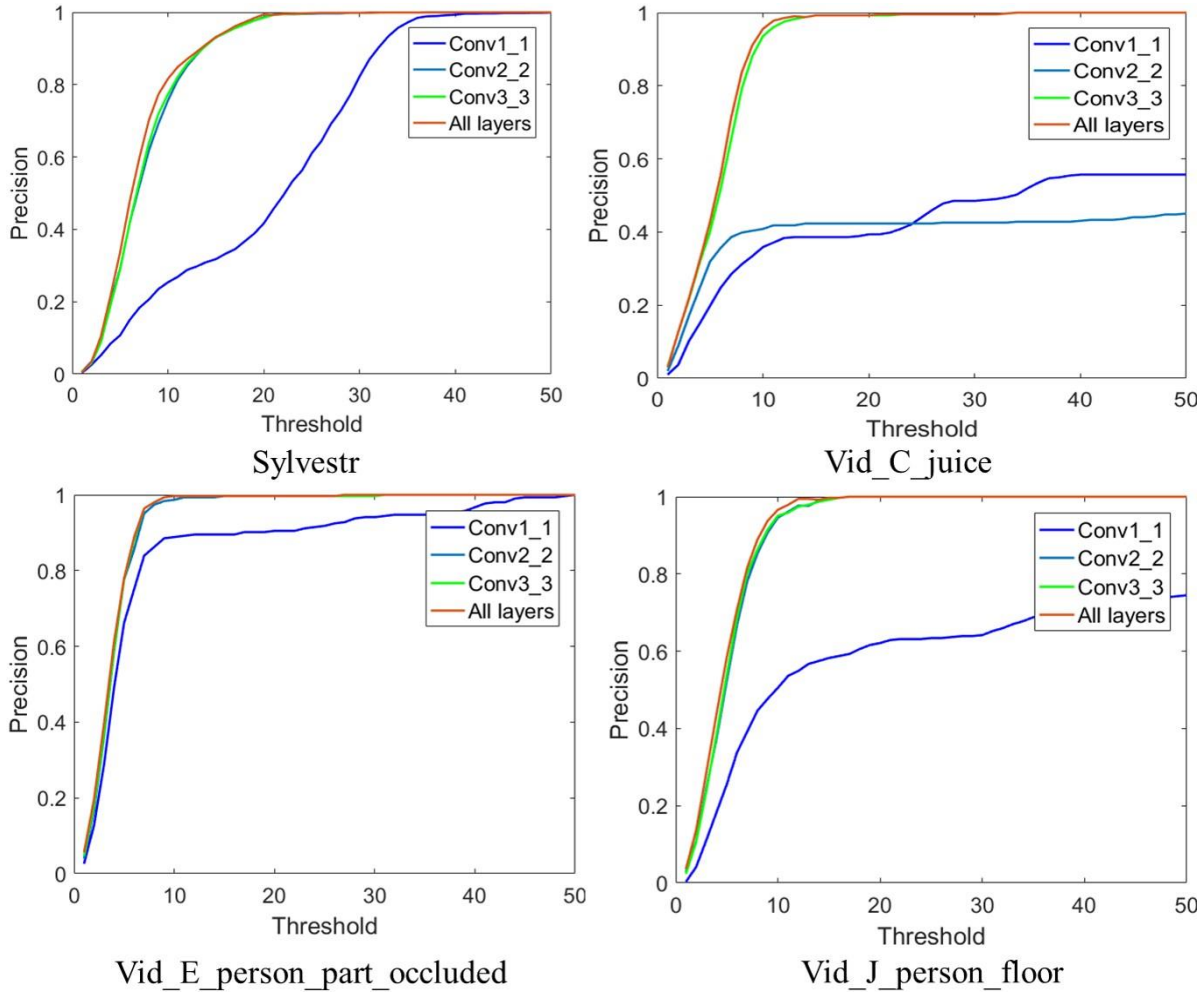

**Fig. 4** Tracking performance with different feature representation from different convolutional layers of the employed CNN.

## 4.2 Robustness to Hyper-parameters

The proposed method requires to specify two hyper-parameters, i.e., the event number in each segment and the convolutional layers for feature representation. To investigate the influence of these two hyper-parameters, we performed a series of experiments on several event-stream recordings.

*Event number in each segment.* We evaluate the influence of the number of events in each segment on several recordings, including the "Horse toy" from DVS128 sensor and the "Sylvestr", "Vid J person floor" from DAVIS sensor. Figure 3 shows the tracking performance of the proposed method under different event number in each segment. The temporal change of the *x* position and *y* position of the center of the tracker are used to show the tracking performance. The number of segments is different from that of the groundtruth under different event number in each segment. It is impossible to compare the position of the tracker in each segment with that in the groundtruth one to one when both have different number of segments. We use the index of the events to represent the time coordinate because different number of events in each segment results in different number of segments. We plot the curve of the *x* position and *y* position of the tracker over the index of events. From the results we can see that the proposed method is robust to the number of events in each segment by comparing the degree of proximity of the trajectory of the tracker and the groundtruth. The effectiveness of the proposed method is owing to the discriminative feature representation transferred from the multiple layers of pre-trained CNN on the computer vision task. Generally, less events are generated in a relatively short time period, and then the proposed method has many potential applications in many the high-speed scenes.

*Feature map from different layers in the CNN.* We test the influence of different convolutional layers of VGG-Net-16 on the tracking performance. We evaluate five kinds of feature representations, i.e., the Conv1 1 convolutional layer, the Conv2 2 layer, the Conv3 3 layer, and the composition of the three convolutional layers. Table 2 shows the spatial size and dimensionality of the feature maps of the input layer and three different convolutional layers. All the DAVIS recordings ("Sylvestr", "Vid B cup", "Vid C juice", "Vid E person part occluded", and "Vid J person floor") are used in this test. For the "Vid B cup" scene, the tracking would fail when a single convolutional layer or a composition of two layers (Conv1 1 or Conv2 2 or Conv3 3) is used. This is because in the "Vid B cup" scene, the target cup is moved over a complicated background while the events from target object are very easy to be mixed up with the events from the background. Figure 4 shows the metric results on the rest four event-stream recordings with different convolutional layers. Feature representation from the higher convolutional layers results in better tracking performance. For the more complicated scenes with complex background, such as the "Vid B cup", combination of hierarchical feature representation from multiple convolutional layers is required for effective object tracking. This is because the feature representation from multiple convolutional layers combines the low-level texture features and high-level semantic features and can handle the fast changing of the appearance of the target object. While for the relatively simple scenes with less noise events and simple background textures, less and lower convolutional layers result in effective tracking with a higher speed.

## 4.3 Tracking speed under different layers

In this section, we investigate the tracking speed of the proposed method on several recordings. The speed is measured over different convolutional layers of the employed CNN net. Table 3 shows the results of tracking speed on five DAVIS recordings. We measure the tracking speed using the unit of segments per second which is similar to the frame per second in the computer vision tracking. In the experiment setup, a PC machine with Intel(R) Core(TM)i5-7300HQ CPU @ 2.5 GHz is used. We did not present the measurement result on some convolutional layers in some event-stream recordings (such as the Conv1 1, and Conv2 2 in the "Vid C ﹍ juice" scene) which have failed in the tracking of the corresponding target object.

Intuitively, high-level layers require more computational operations, which results in the decrease of the tracking speed. Then the tracking precision and the speed are a tradeoff. In the simple scenes, such as monitoring an object with a fixed sensor, low-level convolutional layers can be used for effectively tracking with high speed. In other complicated scenes such as complicated background textures or moving DVS sensor, high level convolutional layers should be used for correct tracking, which decrease the tracking speed. The relatively high speed of the proposed method even with high-level feature representation is owing to the high computational efficiency of the correlative filter mechanism in the frequency domain.

## 4.4 Comparison with Other Methods

In this section, we compare the performance of the proposed method to some other event-stream pattern tracking methods. We perform the experiments on all the eight event-stream recordings. We introduced the four other tracking methods as follows,

*Three tracking methods in jAER software.* These three methods are based on the three event filters in jAER source available within the jAER sourceforge repository, including "Rectangular Cluster Tracker", "Einstein Tracker", and "Median Tracker", respectively. Since the three methods have failed in the more complicated scenes, we only provide the tracking results of the three methods on the three simple scenes captured by DVS128 sensor.

*Compressive tracking based rate coding feature.* This is a feature based event-stream tracking method based on compressive sensing and has achieved good performance on some simple scenes captured by DVS128. The compressive tracking learns a classifier on the compressive coding of haar-like feature extracted on the rate coding map. We provide the results of this method on all the event-stream recordings for comparison with the proposed method.

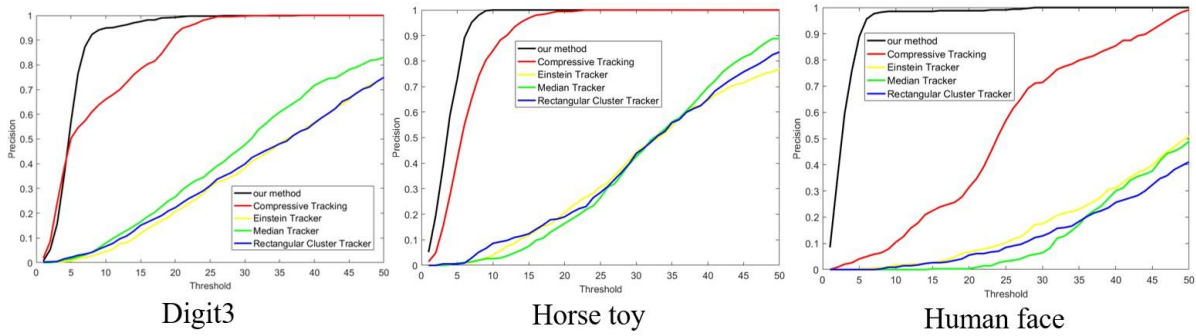

**Fig. 5** Tracking results on three event-stream recordings captured by DVA128 sensor (from left to right are Digit3, Horse toy, and Human face, respectively).Comparison with four other tracking methods.

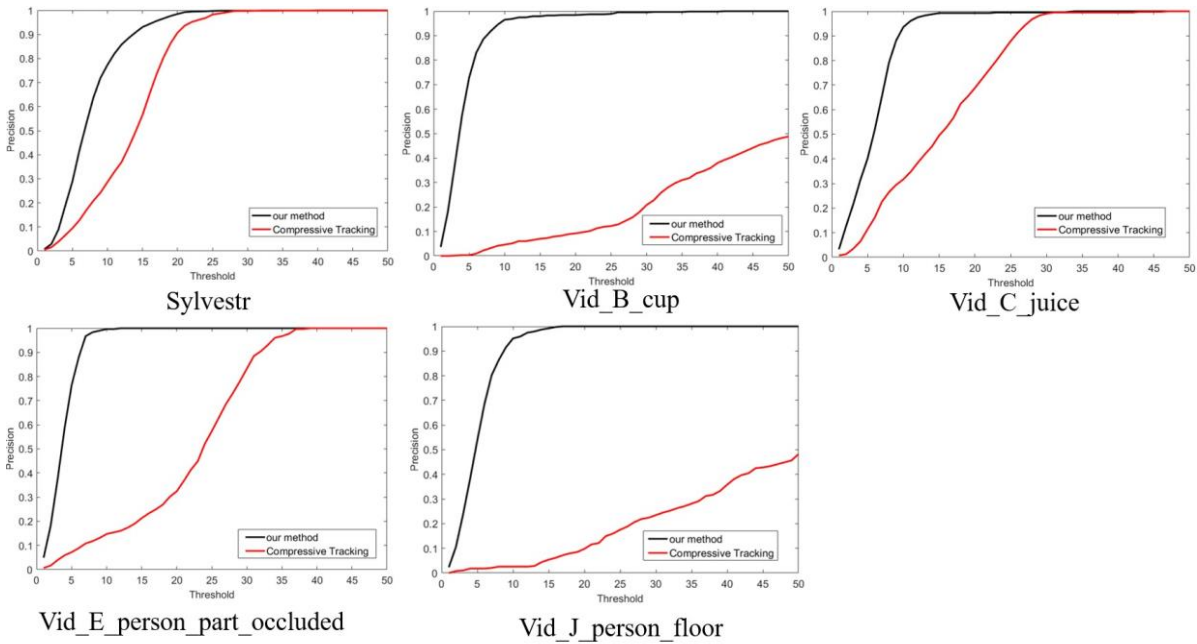

**Fig. 6** Tracking results on five event-stream recordings captured by DAVIS sensor (from left to right are Digit3, Horse toy, and Human face, respectively).Comparison with the compressive sensing based tracking methods.

Figure 5 shows the comparison result of the proposed method with all the four methods on the DVS128 recordings. Center location error is used to measure the tracking performance. Results show that the proposed tracking method achieves the best performance. Three tracking methods in jAER software achieved the bad performance because the three DVS recordings contains some challenges include noise events, fast deformation, and fast changing pose. Compared to the compressive tracking method, the proposed method show better performance for the more discriminative CNN feature presentation than the haar-like feature in the compressive tracking.

On the five DAVIS recordings, we only compared the tracking performance of the proposed method with the compressive tracking method because the three tracking methods in jAER software fail to track the target object in more complicated scenes. Figure 6 shows the tracking location precision of the two method. The proposed method achieved better performance on the five DAVIS recordings. Especially, in the "Vid B cup" scene where many objects in the background have the same shape with the target object "cup", the proposed tracker can track the target object correctly. Other feature based methods with simple feature representation of the changing target appearance cannot handle many complicated scenes with noise events, and complicated background textures, fast changing appearance, and occlusion. By combining the low-level texture features and high-level semantic features, the feature representation from multiple convolutional layers handles the complicated scenes very well. Results demonstrate that the proposed method is more robust to many challenging visual scenes with better tracking performance than other methods. .

We also display some tracking example by integrating the events into reconstructed frames as shown in Figure 7. In the tracking process, we did not change the scale of the tracker. The proposed tracker tracked the target object with high location precision while the other compressing sensing based tracker drifts very easily. Event with complicated background and occlusion, our tracker show very high tracking precision.

## 5 Conclusion

In this work, we proposed a robust event-stream object tracking method based on the CF tracking mechanism. Our method overcomes many challenges in the event-stream tracking, such as the noise events, chaos of the complex background texture, occlusion, and randomness of event generating in the pixel circuit. Rate coding is used to encode the visual information of the target event-stream object. Correlation response map is computed on the feature representation from the hierarchical convolutional layers of a pre-trained deep CNN.

The proposed method shows good performancein many complicated visual scenes. Compared with other feature based methods, our method is more robust in many visual scenes with noise and complicated background textures. Compared with other event-driven method, the proposed method has real-time advantage on event streams with large number of events. To utilize each single event for tracking and updating the appearance of the target object without segment reconstruction is a more interesting and challenging task which has lied in the heart of current event-based vision research and will be explored in the future. In event-driven tracking task, how to suppress the noise events is an import step for correct tracking as the noise events will lead the tracker to make wrong estimation.

**Acknowledgements** The work was partially supported by the Project of NSFC (No. 61327902), Beijing program on study of functional chip and related core technologies of ten million class of brain inspired computing system (Z151100000915071) and Study of Brain-Inspired Computing System of Tsinghua University program (20141080934, 20151080467).

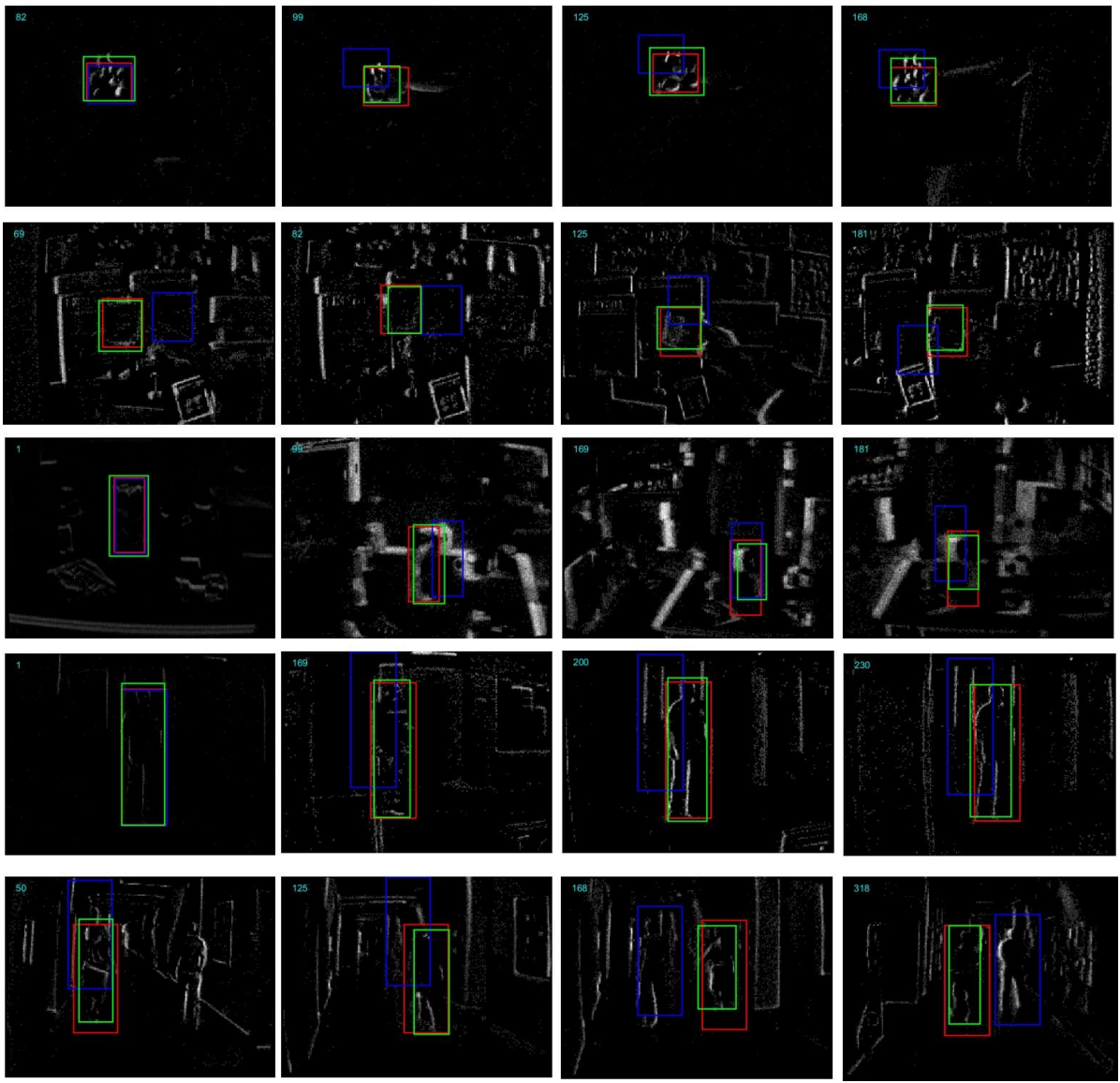

**Fig. 7** Comparison of the proposed tracking method with the compressive sensing based method on example segments from the five DAVIS recordings. From top to bottom are the "Sylvestr", "Vid B cup", "Vid C juice", "Vid E person part occluded", and "Vid J person floor", respectively. The bounding boxes with green, red, and blue color represent the groundtruth, our proposed tracker and the compressive sensing based tracker, respectively.

**Table 1** Challenges of each recording. If the recording (in the row) has this challenge (in the column), then the corresponding value is set to 1, otherwise 0.

| Challenge | Noise events | Complicated background | Occlusion | Intersected trajectories | Deformation | Scale change | Pose change |
|---|---|---|---|---|---|---|---|
| Digit3 | 1 | 0 | 0 | 0 | 0 | 1 | 0 |
| Horse toy | 1 | 0 | 0 | 0 | 1 | 1 | 1 |
| Human face | 1 | 0 | 0 | 0 | 1 | 1 | 1 |
| Sylvestr | 1 | 0 | 0 | 0 | 1 | 1 | 1 |
| Vid B cup | 1 | 1 | 0 | 0 | 0 | 0 | 1 |
| Vid C juice | 1 | 1 | 0 | 1 | 0 | 1 | 0 |
| Vid E person part occluded | 1 | 1 | 1 | 0 | 0 | 0 | 0 |
| Vid J person floor | 1 | 1 | 1 | 1 | 1 | 1 | 1 |

**Table 2** The spatial size and dimensionality of the feature representations from three different convolutional layers of the employed network. Input layer denotes the input rate coding map, after the necessary preprocessing steps.

|  | Input | Conv1_1 | Conv2_2 | Conv3_3 |
|---|---|---|---|---|
| Spatial size | M*N | M*N | M/2*N/2 | M/4*N/4 |
| Dimensionality | 3 | 64 | 128 | 256 |

**Table 3** The tracking speed under different convolutional layers of the employed network. The tracking speed is measured by the segments per second. We measure the tracking speed on the DAVIS recordings.

| Recording | Conv1 1 | Conv2 2 | Conv3 3 | All |
|---|---|---|---|---|
| Sylvestr | 114.1 | 68.8 | 38.2 | 30.1 |
| Vid C juice | - | - | 37.8 | 29.8 |
| Vid E person part occluded | - | 68.5 | 37.9 | 29.2 |
| Vid J person floor | - | 68.3 | 38.1 | 29.0 |
| Vid B cup | - | - | - | 29.3 |